\documentclass[journal]{IEEEtran}

\ifCLASSINFOpdf
\else
\fi
\usepackage{graphicx}
\usepackage{amsmath}
\usepackage{amsthm}
\usepackage{booktabs}
\usepackage{algorithm}
\usepackage{algorithmic}
\usepackage{amssymb}
\usepackage{bbm} 
\usepackage{multirow}
\usepackage{subfigure} 
\usepackage{bm}

\begin{document}

    \title{NI-UDA: Graph Adversarial Domain Adaptation from Non-shared-and-Imbalanced Big Data to Small Imbalanced Applications}

 \author{Guangyi~Xiao,~\IEEEmembership{Member,~IEEE,}
		Weiwei~Xiang,~
	Huan~Liu,~
	Hao~Chen,~
	Shun~Peng,~
	Jingzhi~Guo,~\IEEEmembership{Member,~IEEE,}
	Zhiguo~Gong,~\IEEEmembership{Senior~Member,~IEEE}
	\thanks{G. Xiao, H. Liu, H. Chen, S. Peng are with the College of Computer Science and Electronic Engineering, Hunan University, Chansha, China 410082 E-mail:guangyi.xiao@gmail.com, hliu@hnu.edu.cn, chenhao@hnu.edu.cn, psbetter@hnu.edu.cn}
	\thanks{W. Xiang is with Key Laboratory of Intelligent Control Technology for Wuling-Mountain Ecological Agriculture in Hunan Province, School of Electrical and Information Engineering, Huaihua University, Huaihua, Hunan 418008, China E-mail: mr\_menand@126.com}
	\thanks{J. Guo, is with the Department of Computer and Information Science, University of Macau, Macau SAR 999078, China E-mail:jzguo@umac.mo}
	\thanks{Z. Gong, is with the State Key Laboratory of Internet of Things for Smart City, and the Department of Computer and Information Science, University of Macau, Macau SAR 999078, China E-mail:fstzgg@umac.mo}
}


\maketitle

\begin{abstract}
We propose a new general Graph Adversarial Domain Adaptation (GADA) based on semantic knowledge reasoning of class structure for solving the problem of unsupervised domain adaptation (UDA) from the big data with non-shared and imbalanced classes to specified small and imbalanced applications (NI-UDA), where non-shared classes mean the label space out of the target domain. Our goal is to leverage priori hierarchy knowledge to enhance domain adversarial aligned feature representation with graph reasoning. In this paper, to address two challenges in NI-UDA, we equip adversarial domain adaptation with Hierarchy Graph Reasoning (HGR) layer and the Source Classifier Filter (SCF). For sparse classes transfer challenge, our HGR layer can aggregate local feature to hierarchy graph nodes by node prediction and enhance domain adversarial aligned feature with hierarchy graph reasoning for sparse classes. Our HGR contributes to learn direct semantic patterns for sparse classes by hierarchy attention in self-attention, non-linear mapping and graph normalization. our SCF is proposed for the challenge of knowledge sharing from non-shared data without negative transfer effect by filtering low-confidence non-shared data in HGR layer.  Experiments on two benchmark datasets show our GADA methods consistently improve the state-of-the-art adversarial UDA algorithms, e.g. GADA(HGR) can greatly improve f1 of the MDD by \textbf{7.19\%} and GVB-GD by \textbf{7.89\%} respectively on imbalanced source task in Meal300 dataset. The code is available at https://gadatransfer.wixsite.com/gada.
\end{abstract}

\begin{IEEEkeywords}
Adversarial domain adaptation, Hierarchical graph reasoning, Non-shared-and-Imbalanced UDA, Big data transfer learning.
\end{IEEEkeywords}

\IEEEpeerreviewmaketitle

\section{Introduction}
The technique of Unsupervised Domain Adaptation (UDA)\cite{zhuang2019comprehensive} is popularly used for (1) handling the scarcity of labelled data in a target domain by leveraging rich labelled data from a related source domain; (2) reducing the cost of training downstream models by enabling models to be reused on other networks.
Recent studies in deep learning reveal that deep networks can disentangle explanatory factors of variations behind domains\cite{yosinski2014transferable,he2016deep}, thus learn more transferable features to improve UDA significantly. A major line of the existing UDA methods formally reduce domain shift and bridge the gap cross-domains by learning domain-invariant feature representations. Generally, these deep domain adaptation methods can be typically categorized into three major categories, including discrepancy-based\cite{ding2018graph}, adversarial-based \cite{cui2020gradually}, and parameter-based domain adaptation\cite{chang2019domain}.

\begin{figure}[t]
	\begin{minipage}[b]{1\linewidth} 
		\centering
		\includegraphics[width=2.8in]{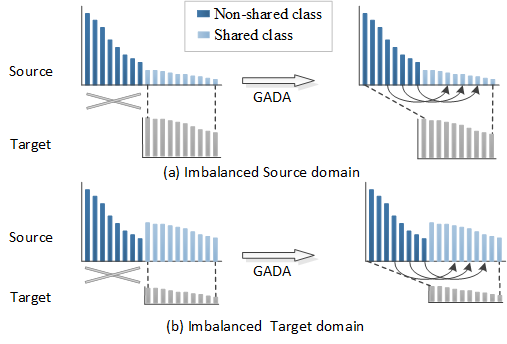}
		\caption{Non-shared and Imbalanced UDA. Left parts are the traditional UDA with ignoring the non-shared data and the traditional PDA with removing the non-shared data automatically, respectively. Right parts are using non-shared data and semantic graph reasoning to address NI-UDA.}
		\label{problem}
	\end{minipage}%
	\hfill  

\end{figure}


How to effectively use big data information to transfer to the target specified small sample area to improve the prediction effect is a key issue for reliable big data transfer learning. UDA mainly realizes the inter-domain difference alignment of a specific shared classes by learning the invariant feature representation between domains. However, big data has the characteristics of large scale, diverse types and imbalances, high noise and rapid growth. Traditional UDA either ignores \cite{cao2019learning} or automatically deletes non-shared big data\cite{long2018conditional,Zhang2019}, which can no longer meet the requirements of effective supervision of data utilization and adaptation of massive complex data in the era of big data. As more and more intelligent decision-making tasks of smart devices and services are subdivided by the industry, the problem of the difference between the distribution of source big data with large-scale non-shared and imbalanced classes and the distribution of subdivision task data with specified small-and-imbalanced classes has become increasingly prominent, which has become an urgent need one of the research problem to be solved \cite{Xiao2020}.

For non-shared and imbalanced big data and small application tasks in domain adaptation, we mainly analyze the major needs of two typical industrial requirements.
(1) Catering big data analysis \cite{Xiao2020}, such as domain adaptation from food material big data to small tasks of restaurant identification. The task of automatic identification of food categories by electronic scales has main challenges: 1) transfer from the laboratory to the real environment, and 2) imbalance of food data. (2) Legal big data analysis \footnote{https://wenshu.court.gov.cn/}, such as the domain adaptation from professional big data to the Q\&A community to automatically judge prediction tasks. Main challenges: 1) Imbalanced professional fact description data. In the professional fact description data set, the distribution of various crimes and legal provisions is very uneven. In the crime distribution, the coverage rate of the top 10 most frequent crimes is as high as 78.1\%, while the coverage rate of the last 50 most frequent crimes is only about 0.5\%. 2) Non-shared imbalanced big data is adapted to some imbalanced target domains.

This paper models the domain adaptation process from big data to small tasks, and is committed to simultaneously achieving big data hierarchy knowledge generalization, non-shared and imbalanced domain adaptation under the imbalance of the source domain and the target domain, so as to provide reliable, efficient, generalized, flexible, and scalable intelligent decision-making for large-scale subdivided smart devices and services for big data analysis. The challenges faced stem from two scientific questions: (1) How to construct non-shared and imbalanced reliable domain adaptation for the non-shared and imbalanced characteristics of source big data. (2) How to transfer from non-shared imbalanced big data to small imbalanced target domain to achieve reliable domain adaptation.  We aim to tackle the problem in domain adaptation process from big data to small tasks, and model it as \textbf{Non-shared-and-Imbalanced UDA (NI-UDA)}.


The challenges of NI-UDA are: (1) \textbf{Knowledge sharing from non-shared data without negative transfer effect}: How to maximize the aggregation of the structural information brought by these similar non-shared classes while minimizing the negative transfer effect\cite{cao2019learning} caused by these dissimilar non-shared classes? (2) \textbf{Sparse classes transfer}: Sparse shared classes in either domain are hard to produce effect in aligning with their partner classes in UDA\cite{Zhang2019}. Since the small size of sparse classes, the aligning effect of the sparse classes will be ignored in the UDA. For example, 10 images of "wildcat" in target domain will be ignored in UDA, which will led to huge negative false (NF) and negative positive (NP) to the sparse "cat" class for the target prediction.



In this paper, we propose \textbf{GADA} to address the challenges of NI-UDA in adversarial domain adaptation manner, which improve the alignment of adversarial domain adaptation with Hierarchy Graph Reasoning (HGR) and Source Confidence Filter (SCF) to achieve jointly knowledge sharing between non-shared classes and shared classes, and between different domains. 

The previous work of TAN\cite{Xiao2020} learned from K-classes in the source domain, ignoring the negative transfer effects brought about by these non-shared classes, that is, error accumulation\cite{Chen2019} and uncertainty propagation\cite{liang2020balanced}. In order to solve the challenge of knowledge sharing from non-shared data without negative transfer effect, we propose a \textbf{Source Classifier Filter (SCF)} mechanism to filter non-shared data with a fixed confidence threshold of source classifier in feature enhance layer. For domain adaptation, SCF can filter 80\% low confidence non-shared data in the initial domain adaptation stage as well as 30\% non-shared data in the converge domain adaptation stage.

The \textbf{motivation example} of HGR to address the challenge of sparse classes transfer is: given the image of sparse "wildcat", suppose the feature representation of "wildcat" is predicted to "wild dog" and "wildcat" with ambiguity scores; then with hierarchy graph reasoning, the feature of sparse "wildcat" will be enhanced by "wildcat"’s parent and sibling nodes (rich "house-cat") or "wild dog"’s parent and  sibling nodes ("house dog"), respectively; since the feature representation is more similar to the pattern of "house-cat", the feature representation is enhanced by rich "house-cat", finally initial ambiguity decision of sparse "wildcat" is definitely classified to "wildcat" class. Meanwhile, once the image of rich "house-cat"is well classified, our HGR can speed up the learning for the semantic pattern of its parent and sibling nodes (sparse "wildcat") directly. 

In HGR, the prior hierarchy of big source labels is easy to construct based on WordNet\cite{miller1998wordnet} or Wikipedia or Knowledge Graph or human experts. Once the hierarchy of labels is constructed, we can leverage priori hierarchy knowledge to enhance domain adversarial aligned feature representation for sparse classes by the hierarchy graph reasoning of big source labels. 
For example, the sparse target "wildcat" can align with the source "wildcat" by graph reasoning of "wildcat" and its sibling rich "house cat". 
However it has a difficult to learn semantic patterns on adversarial domain adaptation for sparse classes.
Our HGR has following new characters: (1) our HGR learns semantic patterns directly by the node predictions of sparse classes in self-attention;
(2) mapping the local feature space of sparse classes and semantic graph space non-linearly; (3) adding normalization to graph reason for transferable feature enhancement. 

The main contributions are: (1) A novel general framework GADA with HGR and SCF mechanism is proposed to address NI-UDA challenges in imbalanced source scenario and imbalanced target scenario. (2) The key component HGR layer could enhance domain adversarial aligned feature representation of sparse classes by effect and direct semantic pattern learning. (3) GADA consistently improves the SOTA of adversarial UDA on multiple datasets under NI-UDA setting. 
\begin{figure*}[t]   
	\begin{minipage}[b]{1\linewidth} 
		\centering   
		\includegraphics[width=7in]{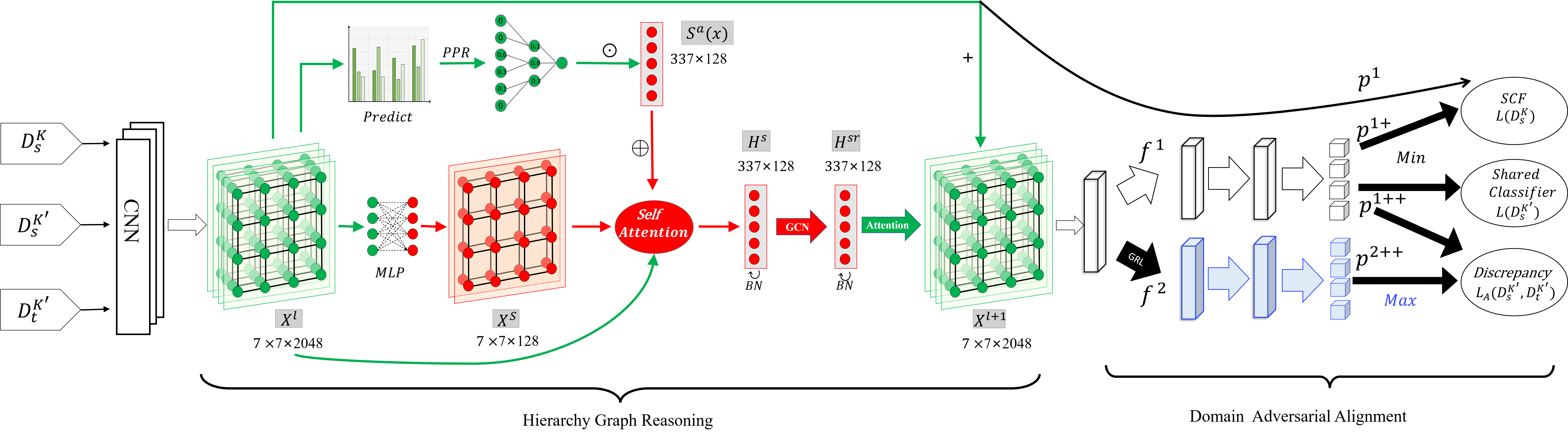}  
		\caption{The work-flow of GADA framework to solve two challenges of NI-UDA with Hierarchy Graph Reasoning (HGR) and Source Classifier Filter (SCF). HGR enhances domain adversarial aligned feature representation of sparse classes in hierarchy graph reasoning: (1) self-attention with hierarchy attention of prediction, (2) non-linearly mapping of two spaces with MLP, (3) normalized graph reasoning for transferable feature enhancement. SCF is to filter non-shared data with a fixed confidence threshold of source classifier for reducing negative transfer effect in HGR layer.}
		\label{figure3}
	\end{minipage}%
	\hfill  
\end{figure*} 
\section{Related works}
\textbf{UDA.} UDA can be mainly categorized into three main groups\cite{zhuang2019comprehensive}: (1) Discrepancy-based domain adaptation leverages different criterion to  minimize the domain discrepancy, like class criterion\cite{yan2017mind}, architecture  criterion\cite{li2018adaptive} and statistic criterion\cite{Zhang2019,ding2018graph}.  (2) Adversarial-based domain adaptation that introduce a domain discriminator and exploits the idea of adversarial learning to encourage domain confusion in minmax game \cite{jiang2020implicit,long2018conditional,cui2020gradually}. A domain classifier is trained to tell whether the sample comes from source domain or target domain. The feature extractor is trained to minimize the classification loss and maximize the domain confusion loss. (3) Parameter-based domain adaptation\cite{chang2019domain,li2018adaptive} assumes that models of the source and target domain share the same prior parameters.
 Some other UDA generalization paradigms are proposed\cite{liang2020we}.\par

\textbf{PDA, ODA and Label Shift.} PDA consider as setting where the label of target domain is a subspace of the source domain with the hard assumption of non-shared classes space is \textbf{unknown}. So these PDA methods are proposed to estimate the non-shared classes space and try to remove the examples in non-shared classes automatically\cite{cao2018pada,cao2019learning,liang2020balanced}. 
Different from PDA, Open-Set Domain Adaptation(ODA) \cite{panareda2017open} assumes that target domain has some unknown non-shared classes. Our NI-UDA assume the non-shared classes space is \textbf{known}, but it still has big challenges on transfer learning on sparse shared classes in either domain and how to borrow knowledge from non-shared classes while without the negative transfer effect. Label shift\cite{zhao2019learning} is more challenging than traditional UDA and common for real-world application\cite{Xiao2020}. 
The potential label shift, which cause conditional feature distributions alignment become more difficult\cite{tan2019generalized}. 
Our NI-UDA can be classified as a special label shift where the distribution of the non-shared classes in target domain is empty.

\textbf{GNNs.} GNNs\cite{zhou2020graph} perform a message passing strategy, where each node update the representation vector by recursively aggregating vectors of neighbor nodes. Recently, GCN\cite{kipf2016semi} and its variants\cite{chien2021adaptive} have emerged in different tasks, demonstrating the ability to capture the inner structure. Liang et al.\cite{liang2018symbolic} performs reasoning over a group of symbolic nodes whose outputs explicitly represent different properties of each semantic in a prior knowledge graph. Ma et al.\cite{Ma2019} proposed GCAN to perform structure-aware alignment by aggregating data structure knowledge for unsupervised domain adaptation. Wang et al.\cite{wang2020learning} leverages the knowledges learned from multiple source domain by combing global prototypes and query samples.

\section{Graph Adversarial Domain Adaptation}

\subsection{Notations and Definitions} 
In NI-UDA, the big data domain covers more classes with more training data in each of them. Let $C^K$ and $C^{K'}$ denote the big data label space and specified small shared class spaces, respectively, so we have $C^{K'}\subseteq C^K$ and $K'\ll{K}$ in general. We denote  $C^{K'}$ in source domain of big data as shared classes, and $C^K-C^{K'}$ as non-shared classes in big data. Suppose $D^K_s=\{x_s,y_s\}$ and $D^{K'}_s=\{x_s,y_s\}$ represent the training data with respect to $C_s^K$ and $C_s^{K'}$ respectively, in the source domain, and  $D^{K'}_t=\{x_t\}$ denotes the training data in the small target domain with respect to  $C_t^{K'}$, where $x_s$ and $x_t$ represent the training data while labels $y_s$ are the corresponding categories of $x_s$. So, we also have $D^{K'}_s\subseteq{D^K_s}$. We have two imbalanced scene settings: (1) \textbf{Imbalanced Source Setting}: the data of source domain shared classes is sparsely distributed; (2) \textbf{Imbalanced Target Setting}: The target domain shared classes data is unlabelled and sparsely distributed.

The source and target domains are drawn from different probability distributions $P$ and $Q$ respectively. Beside $ P(D_s^K) \neq Q(D_t^{K'}) $, we further have $P(D_s^{K'}) \neq  Q(D_t^{K'})$  in our NI-UDA, where $P(D_s^{K'})$ denotes the distribution of shared-class data in source domains.  For such a situation, we want to build a robust deep learning solution to transferring knowledge from $D^K_s$ to $D^{K'}_t$, where $D^K_s\rightarrow{D^{K'}_t}$ is called \textbf{NI-UDA}. 

\subsection{GADA method}
\textbf{NI-UDA} problem is addressed by equip the domain adversarial alignment with HGR for sparse classes transfer and SCF for knowledge sharing from non-shared data without negative transfer effect, respectively. The architecture of our proposed GADA is shown in Fig. 2. We train a K-classes classier $f(x)$ to learn discriminative and domain invariant features $HGR(CNN(x))$ by minimizing the overall objective:
\begin{small}
	\begin{equation}
	\begin{split}
	\mathcal{L}(D_s^K,D_s^{K'},D_t^{K'}) = \mathcal{L}_{K'}(D_s^{K'}) \\ + {\lambda}_1  \mathcal{L}_K(D_s^K) 
+  {\lambda}_2 \mathcal{L_{A}}(D_s^{K'},D_t^{K'})
    \end{split}	\label{1}
	\end{equation}
\end{small}where ${\lambda}_1, \lambda_2$ are the trade-off factors parameters, the first item is the cross-entropy loss for shared source domain, the second item is the cross-entropy loss for non-shared data with SCF, the last item is the adversarial domain discrepancy between shared source domain and the specified small target domain.

In network configuration, similar to adversarial domain adaptation, we not only use a backbone CNN network $CNN(x)$, two classifier networks $f^1$ and $f^2$ in MDD \cite{Zhang2019}, but also use a feature enhance layer $HGR(CNN(x))$ for sparse classes. The whole GADA network configuration are denoted as:
\begin{small}
\begin{equation}
\begin{split}
p^{1}(x) = \sigma (f^1(CNN(x))), \\
p^{1+}(x) = \sigma (f^1(HGR(CNN(x)))), \\
P^{1++}(x) = p^{1+}(x) \cdot \mathbf{m}, \\
h^1(x) = argmax(p^{1++}); \\
p^{2+}(x) = \sigma (f^2(HGR(CNN(x)))), \\
P^{2++}(x) = p^{2+}(x) \cdot \mathbf{m}, \\
h^2(x) = argmax(p^{2++})
\end{split} \label{2}
\end{equation}
\end{small}where $\mathbf{m}$ is the mask of the one hot representation of $K'$ large-scale and flexible specified target shared classes with all K-classes in the big data domain, $\sigma$ is the softmax operation, and $h^1(x)$ is the argmax operation. Therefore, $p^{1}(x)$ is the K-classes prediction score with backbone feature $CNN(x)$ and the $f^1$ classifier; $p^{1+}(x)$ is the enhanced K-classes prediction score with backbone feature $CNN(x)$, feature enhance layer $HGR(CNN(x))$ for sparse classes and the $f^1$ classifier; to focus on the prediction of specified small target classes, we use the flexible mask $\mathbf{m}$ times $p^{1+}(x)$ to form our target predict score $P^{1++}(x)$; at last, we can use argmax operation $h^1(x)$ to produce the predict label. Note that, we denote $p^{1+}(x)$ and $p^{2+}(x)$ since them use different classifiers $f^1$ and $f^2$ respectively.  

\textbf{Small Shared Classifier.} Similar to the source classifier loss in UDA, the classification loss of $K'$ of shared classes is defined as:
\begin{small}
\begin{equation}
\begin{split}
\mathcal{L}_{K'}(D_s^{K'})=\mathbb{E}_{D_s^{K'}} [-log(p^{1++}_{y^s}(x^s)]\label{3}
\end{split}
\end{equation}
\end{small}where $D_s^{K'}$ is the shared source dataset, $p^{1++}_{y^s}(x^s)$ is the target predict score for the example $(x^s, y^s)$, and the cross-entropy loss is used for small shared classifier with feature enhance layer $HGR(CNN(x))$ and the flexible mask $\mathbf{m}$.

\textbf{Big Source Classifier with SCF.} To address the challenge of knowledge sharing from non-shared data without negative transfer effect in NI-UDA, we propose the Source Classifier Filter (SCF) mechanism in big source classifier. The previous work of TAN learned from K-classes in the source domain, ignoring the negative transfer effects brought about by these non-shared classes, that is, error accumulation\cite{Chen2019} and uncertainty propagation\cite{liang2020balanced}. In order to solve this problem, we propose a \textbf{SCF} mechanism. In the backbone feature $CNN(x)$ learning, non-shared classes should be fully learned, thus we have the loss:
\begin{small}
\begin{equation}
\mathcal{L}_K^1(D_s^K)=\mathbb{E}_{D_s^{K}} [-log(p^{1}_{y^s}(x^s)] \label{4}
\end{equation}
\end{small}where $D_s^{K}$ is the big source dataset. However in the enhanced feature learning, hierarchical graph inference will mistakenly learn low-confidence samples, leading to error accumulation. So we propose Sourse Classifier Filter to maximize the aggregation of the structural information while minimize the negative transfer effect in HGR layer. Thus in the feature enhance layer $HGR(CNN(x))$ learning, the enhanced feature learning loss with \textbf{SCF} is defined as:
\begin{small}
\begin{equation}
\begin{split}
\mathcal{L}_K^2(D_s^K)=\mathbb{E}_{D_s^{K}} [-log(p^{1+}_{y^s}(x^s))]\mathbbm{1}(p^1_{y^s}(x^s) > \gamma)  
\label{5}
\end{split}
\end{equation}
\end{small}where $\gamma$ is the confidence filter threshold, and $p^{1+}(x)$ is the predictor with enhanced feature $HGR(CNN(x))$, in compare with the origin predictor $p^{1}(x)$ with CNN feature $CNN(x)$. For domain adaptation, SCF can filter 80\% low confidence non-shared data in the initial domain adaptation stage as well as 30\% non-shared data in the converge domain adaptation stage.

 The total big source classifier loss is defined as:
 $\mathcal{L}_K = \mathcal{L}_K^1 + \mathcal{L}_K^2$. Note that both small shared classifier and big source classifier shared the same classifier network $f^1$, the difference and flexibility is achieved by the flexible mask $\mathbf{m}$.


\textbf{Adversarial Domain Discrepancy.} In the pioneer work\cite{ben2010theory}, the $\mathcal{H}\Delta \mathcal{H}$-divergence of two classifies $f^1$ and $f^2$ is used to measure the distribution discrepancy of two datasets. Thus the adversarial domain discrepancy in GADA is integrated as :
\begin{small}
\begin{equation}
\begin{split}
\mathcal{L_{A}}(D_s^{K'},D_t^{K'})= \sup\limits_{h^1,h^2 \in \mathcal{H}}| \mathbb{E}_{D_t^{K'}} \mathbbm{1}[h^1 \neq h^2 ] - \mathbb{E}_{D_s^{K'}} \mathbbm{1}[h^1 \neq h^2 ]| \label{6}
\end{split}
\end{equation}
\end{small}where $\mathbbm{1}$ of two datasets are the indicator functions to measure the prediction discrepancy of two classifiers $f^1$ and $f^2$ in two datasets $D_s^{K'}$ and $D_t^{K'}$, respectively. Note that both $h^1(x)$ and $h^2(x)$ are based on enhanced feature layer $HGR(CNN(x))$ and the flexible mask $\mathbf{m}$.


In state-of-the-art adversarial domain discrepancy\cite{cui2020gvb,Zhang2019},  Marginal Domain Discrepancy (MDD)\cite{Zhang2019} is integrated as:
\begin{small}
\begin{equation}
\begin{split}
\mathcal{L_{A}^{MDD}}(D_s^{K'},D_t^{K'})= \sup\limits_{f^2 \in \mathcal{F}}  \mathbb{E}_{D_t^{K'}}L'(p^{2++}_{h^1(x^t)}(x^t)) \\
- {\lambda}_3 \mathbb{E}_{D_s^{K'}}L(p^{2++}_{h^1(x^s)}(x^s)),  \\
L'(p^{2++}_{h^1(x^t)}(x^t)) = log[1-(p^{2++}_{h^1(x^t)}(x^t))],\\
L(p^{2++}_{h^1(x^s)}(x^s)) = -log[p^{2++}_{h^1(x^s)}(x^s)],
\end{split} \label{7}
\end{equation} 
\end{small}where $L, L'$ are a disparity for each domain using two different classifiers $f^1$ and $f^2$, the pseudo label with $h^1(x)$, and ${\lambda}_3$ is designed hyper-parameter to balance the generalization and the optimization. Note that both $h^1(x)$ and $p^{2++}(x)$ are based on enhanced feature layer $HGR(CNN(x))$ and the flexible mask $\mathbf{m}$. Except these two components for feature enhance representation for spare classes and the classifier with flexible specified small target task, other components are the same as MDD for domain discrepancy.

\textbf{Hierarchy Graph Reasoning (HGR).} To address the challenge of sparse classes transfer in NI-UDA, our HGR enhances domain adversarial aligned feature representation $HGR(CNN(x))$ through hierarchy graph reasoning for sparse classes: (1) aggregate local feature to semantic node with node prediction in self-attention, (2) maps the local feature space and the semantic space non-linearly with MLP, and (3) adds normalization of the semantic nodes in graph reasoning for transferable feature enhancement.

\subsection{HGR Layer}
\textbf{Graph Construction.}
The key input of HGR layer is the flexible deep hierarchy graph for all the labels of big source domain in priori. For the priori knowledge graph, we assume that the classes have been organized into a flexible deep multi-level hierarchy. There are $K$ leaf nodes corresponding to the $K$ leaf classes, which are organized into $\mathcal{G}=(\mathcal{N},\varepsilon)$, where $\mathcal{N}$ and $\varepsilon$ denote the symbol set and edge set, respectively. In this paper, our priori classes hierarchy graph is constructed by a list of entity classes (e.g. "wildcat", "wild dog"), and its graph edge is a concept of belongings (e.g. "is kind of" or "is part of"). The \textit{K} leaf classes and their super classes consist the symbol set $\mathcal{N}$. The edges are the belongings relation between sub-classes and super-classes. For common-sense hierarchy graph we may follow the WordTree or WordNet\cite{miller1998wordnet}.

\textbf{Hierarchy Attention with Prediction.}
To learn semantic pattern directly for sparse classes, we propose a hierarchy attention mechanism of all semantic nodes with the source classifier prediction by the personalised page rank.

\begin{figure*}[!t]  
	\begin{minipage}[b]{1\linewidth} 
		\centering   
		\includegraphics[width=7in]{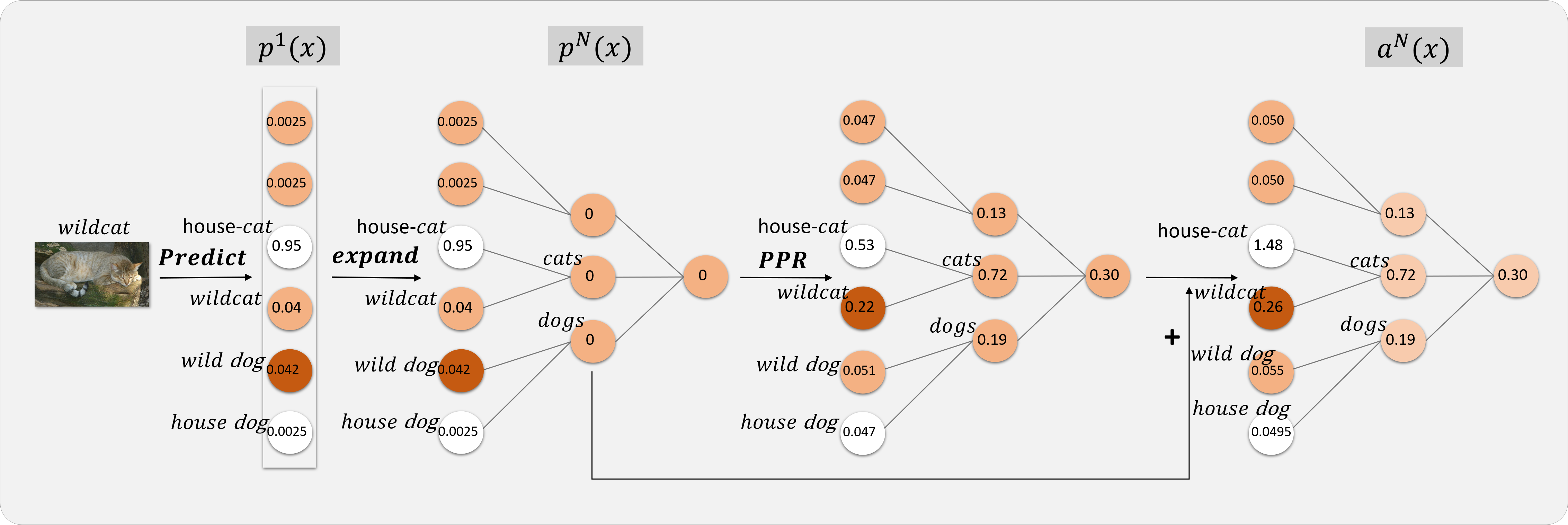}    
		\caption{The workflow of input  $x$(sparse "wildcat") in  Personalized PageRank. To illustrate, we set K=6, N=10, where nodes in White are non-shared classes, sparse "wild-cat" and rich "house-cat" belong to the super-class of cats, and sparse "wild dog" and rich "house dog" belong to the super-class of dogs. Through the node prediction and PPR, initial low score of sparse "wildcat" is increased by the rich "house-cat".}
		\label{ppr}
	\end{minipage}%
	\hfill  
\end{figure*}

Suppose given a input $x$ ("wildcat"), we get the prediction score with  $p^1(x)=\sigma (f^1(CNN(x))) $ where $CNN(X)$ is the CNN feature extractor. Although we get prediction score for leaf nodes for $K$-classes, we have no score for the parent nodes and root node. Therefore personalized page rank algorithm is applied to inference the score for them and enhance the score for sibling nodes. The attention score of graph node sparse "wildcat" will be increased by and its sibling nodes rich "house cat". The workflow is shown in Fig. \ref{ppr}.

We then expand the $K$-classes prediction $p^1(x)$ to $\mathbb{N}$-classes prediction $p^{\mathcal{N}}(x)$ with zero padding which is better than the sum of child nodes in practice. After that, We following the classic personalized Page Rank via power iteration and the normalization operation to calculate the hierarchy attention as:
\begin{small}
	\begin{equation}
	a^{\mathcal{N}}(x) = PPR(p^{\mathcal{N}}(x)) + p^{\mathcal{N}}(x) \label{(8)} 
	\end{equation}
\end{small}where $PPR$ is the personalized page rank algorithm. See example in Fig. \ref{ppr}. Thus, the word embedding of sparse "wildcat" and its sibling nodes rich "house cat" have been attended with the hierarchy attention score. Assume $\mathcal{S}$ is the word embedding of the $\mathcal{N}$ semantic nodes where $\mathcal{S}\in \mathbb{R}^{\mathcal{N} \times D^S}$ and $D^S$ is the designed size of semantic nodes, the attended word embedding of the $\mathcal{N}$ semantic nodes given an input example $x$ is computed by $	S^{a}(x) = a^{\mathcal{N}}(x) \cdot \mathcal{S} $. 

\textbf{Local-to-Graph Self-Attention.}
Meanwhile, given the input $x$ (sparse "wildcat") after all the CNN layers we get local feature tensors $X^l$, our target is to leverage transfer global graph reasoning to form global semantic feature. We aggregate all the local features to each graph node globally with query-key-value attention by the query of attended word embeddings (the word embeddings of sparse "wildcat" and its sibling nodes rich "house cat"). Formally, the feature tensor $X^l\in \mathbb{R}^{H^l\times W^l\times D^l}=CNN(x)$ after $l$-th convolution layer is the module inputs where $H^l$, $W^l$ and $D^l$ are the tensor size. This module aims to produce visual representations $H^{S} \in \mathbb{R}^{|\mathcal{N}| \times D^S}$ of all $|\mathcal{N}|$ symbolic nodes using $X^l$, where $D^S$ is desired feature dimension for each node.

Inspired by \cite{lin2013network}, we transform the local feature space to the semantic space with a two layer MLP as:
\begin{small}
	\begin{equation}
	X^S = MLP(X^l) \label{(9)} 
	\end{equation}
\end{small}where $X^S\in \mathbb{R}^{H^l\times W^l\times D^S}$ and $D^S$ is the designed size of semantic nodes. See ablation study.

\textbf{Self Attention}  is defined as:
\begin{small}
	\begin{equation}
	H^S_n=\sum_{i \in H^l \times W^l} a_{i \rightarrow n} X^S_i, \quad a_{i \rightarrow n} = softmax(W^{aT}_nX^S_i)  \label{(10)} 
	\end{equation}
\end{small}where $X^S_i \in \mathbb{R}^{D^S}$ is the local feature in semantic space, $W^a = \{W^{a}_n\} \in \mathbb{R}^{|\mathcal{N}| \times D^c}$ is the self-attention parameters in local feature space, and the output is $H^S=\{H^S_n\} \in \mathbb{R}^{|\mathcal{N}| \times D^S}$. To add transferability, we normalization graph node with Batch Normalization (BN):
\begin{small}
	\begin{equation}
	H^{s'}=Relu(BN(H^s)) \oplus S^a(x) \label{(11)} 
	\end{equation}
\end{small}where $S^{a}(x)$ is the \textbf{attended word embedding} after prediction of source classifier, $\oplus$ is the concatenate operation, thus the initial attended semantic feature and self-attended local feature are fused in semantic space.\par

\textbf{Semantic Reasoning.}
The visual global graph node is required to be constraint by human priori knowledge, then we evolve the global graph by semantic reasoning. For example sparse "wildcat"'s node feature is constrained by sibling node feature rich "house cat". Formally, the graph reasoning module performs graph propagation over representations of $H^{s'}$ of all symbolic nodes via the matrix multiplication from, resulting in the evolved features $H^{sr}$.

We reused graph convolutional networks \cite{liang2018symbolic} which is defied as:
\begin{small}
	\begin{equation}
	H^{sr}=\mathcal{Q'}^{-\frac{1}{2}} A'^g \mathcal{Q'}^{-\frac{1}{2}} 
	H^{s'} W^g \label{(12)} 
	\end{equation}
\end{small}where $W^g \in \mathbb{R}^{(D^S+D^S)\times D^S}$ is the trained matrix, $A'^g=A^g+I$ is the adjacency matrix of the graph $\mathcal{G}$ with added self-connections for considering its own representation of each node and $I$ is the identity matrix. $\mathcal{Q}'_{ii}=\sum_j A'^g_{ij}$. Using batch normalization to add transferability:
\begin{small}
	\begin{equation}
	H^{sr'}=Relu(BN(H^{sr})) \label{(13)} 
	\end{equation}
\end{small}

\textbf{Graph-to-Local Attention.}
The evolved global representation $H^{sr'}$ of semantic graph nodes (sparse "wildcat" and rich "house cat") can be used to enhance local feature (sparse "wildcat") with attention which is queried by the local features. 

We then transform the semantic feature space back to the local feature space with a two layer MLP as:
\begin{small}
	\begin{equation}
	H^l = MLP(H^{sr'}) \label{(15)} 
	\end{equation}
\end{small}where $H^l\in \mathbb{R}^{\mathcal{N}\times D^l}$ and $D^l$ is the channel size of the local features.

We use the classic query-key-value attention mechanism to aggregate semantic graph nodes in local feature space to the local features as:
\begin{small}
	\begin{equation}
	X^{l'} = Attention(X^l, H^l, H^l) \label{(16)} 
	\end{equation}
\end{small}where $X^l$ is the query, $H^l$ is the key and value, the softmax operation is calculated over the ${\mathcal{N}}$ space.

\textbf{HGR Layer.}
The final output of HGR is the residual of the input local feature $X^l$ and the graph reasoned local feature $X^{l'}$:
\begin{small}
	\begin{equation}
	X^{l+1} = X^{l} + X^{l'} \label{(17)} 
	\end{equation}
\end{small}where $X^l$ is original input of HGR layer. 

Finally, the HGR layer is embedded in adversarial domain adaptation as the enhanced and aligned feature representation $HGR(CNN(x))$ for sparse classes. To constrain the high level feature map with the HGR, so we add one or more HGR layers after the last CNN layer.

\begin{table*}[!t]
	\renewcommand{\arraystretch}{1.0}
	\small
	\caption{Classification accuracy and f1 value (\%) on Meal-300 dataset in Imbalanced Source Setting }
	\label{Tab-meal300-source}
	\centering
	\setlength{\tabcolsep}{1.1mm}{
		\begin{tabular}{c| c c c c c c c c c c c c c}
			\hline
			 & \multirow{3}{*}{Method}
			 & \multicolumn{2}{c}{M300(\textbf{10})} & \multicolumn{2}{c}{M300(\textbf{20})}    & \multicolumn{2}{c}{M300(\textbf{50})}    & \multicolumn{2}{c}{M300(\textbf{70})}    & \multicolumn{2}{c}{M300(\textbf{100})}   & \multirow{3}{*}{Avg-acc}                 & \multirow{3}{*}{Avg-f1}                                                                                                                   \\
			 &                                       & \multicolumn{2}{c}{$\rightarrow$R50(60)} & \multicolumn{2}{c}{$\rightarrow$R50(60)} & \multicolumn{2}{c}{$\rightarrow$R50(60)} & \multicolumn{2}{c}{$\rightarrow$R50(60)} & \multicolumn{2}{c}{$\rightarrow$R50(60)} &                                                                                                                                           \\
			\cline{3-12}

			 &                                       & acc                                      & f1                                       & acc                                      & f1                                       & acc                                      & f1                      & acc              & f1               & acc              & f1               &                                     \\
			\hline

			\multirow{2}{*}{\shortstack{ PDA \\ $300\rightarrow$50}}
			
			 & BA$^{3}$US    & -                                     & -                                     & -                                     & -                                    & 17.55                                    & 7.86                    & 27.28            & 11.72            & 43.89            & 19.04  & 29.57 & 12.87                                                 \\
			 \cline{2-14}
			 & BA$^{3}$US+HGR & - & - & - & -& 19.86 &  8.76 & 28.67 & 12.72 & 46.37 & 20.67 & 31.63 & 14.05  \\ 
			 
			\hline

			\multirow{6}{*}{\shortstack{ UDA \\$ M50 \rightarrow$ \\ R50}}
			 & ResNet               & 45.07                                    & 44.46                                    & 47.91                                    & 47.55                                    & 55.35                                    & 54.68                   & 53.72            & 53.58            & 57.50            & 56.95            & 52.31            & 51.64            \\
			 & CDAN      & 56.31                                    & 55.48                                    & 63.11                                   & 62.87                                    & 69.30                                    & 69.09                   & 70.94         & 70.62           & 71.89            & 71.60            & 66.31            & 65.93          \\
		
			  & GVB-GD & 61.58	& 62.21	 & 68.25 &	68.59 &	74.58	& 74.57	& 75.29	& 75.41	 & 77.20 & 77.21 & 71.38 & 71.58  \\
			  & GCAN & 62.57 & 62.67 &  68.96 &	68.91	&73.52&	73.60 &	75.63 &	75.50 &	76.59	&76.40 &		71.45 &	71.41 \\

			  & MDD                  & 64.13                                    & 64.42                                   & 70.26                                   & 70.15                                    & 73.63                                    & 73.59                   & 74.84            & 74.76               & 75.07            & 74.64    & 71.58             & 71.51           \\
			  \cline{2-14}
			  & MDD+HGR  & 65.87  &	66.09 &	71.24 &	70.39 & 75.36 &	75.29 &	76.26 &	76.11 &	77.08 &	77.06 &	73.16 &	72.98 \\

			\hline

			\multirow{4}{*}{\shortstack{ NI-UDA \\ $M300 \rightarrow$ \\ $M50 \rightarrow R50$ }}

			 & TAN\cite{Xiao2020}                    & 52.88                                    & 52.03                                    & 55.29                                    & 54.94                                    & 59.00                                    & 57.66                   & 60.50            & 59.80            & 61.48            & 61.36            & 57.83            & 57.16            \\
		
		     & GADA(MDD+SGR) & 68.86 & 68.54 & 73.56 &	73.54 &	75.84 &	75.81 & 75.77&75.56 &76.04&75.91 &	74.01&	73.87 \\ 
		     & GADA(MDD+HGR) &  \bm{$71.92$} &	\bm{$71.62$} &	\bm{$74.88$} &	\bm{$74.94$} &	76.47 &	76.14 &	77.13 &	76.94 &	78.18 &	77.95 &	\bm{$75.71$} &	\bm{$75.51$} \\
		     
			  & GADA(GVB+HGR) & 70.33 & 70.10 & 72.77 & 72.62 &	\bm{$77.40$} &\bm{$77.22$} &	\bm{$77.74$} & \bm{$77.57$} &\bm{$79.48$} &\bm{$79.35$} & 75.54 & 75.37 \\

			\hline
		\end{tabular}}
\end{table*}
\section{Experiments}
We conducted experiments on two challenge dataset, where one is the real NI-UDA dataset and the other one is benchmark dataset in UDA.

\subsection{Experimental Settings}

\textbf{Dataset.} \textbf{Meal-300} \cite{Xiao2020} is a challenging lab-to-real dataset of food materials in NI-UDA, with two very distinct domains: \emph{Meal(M)}, 53,374 clear and undisturbed images of 300 classes with long-tail data distribution. \emph{Real(R)}, images are collected in the real electronic receiving environment,it has 50 shared classes, each with 60 images.

\textbf{Office-Home}\cite{venkateswara2017deep} consists of 4 different domains: Art(Ar) with 2,427 images, Clipart(Cl) with 4,365 images, Product(Pr) with 4,439 images, and Real-World (Rw) with 4,357 images. Each domain has 65 classes. For NI-UDA, different from  \cite{cao2019learning}, the Ar domain has fewer and imbalanced images, thus we choose the top 25 classes with \textbf{the least size} the Ar domain as the shared classes to set up imbalanced scenario. Thus, we form three imbalanced source transfer tasks: $Ar \to (Cl, Pr, Rw)$ and three imbalanced target transfer tasks: $(Cl, Pr, Rw) \to Ar$. 

\textbf{Evaluation Protocols.}  Similar to \cite{Xiao2020}, we have three different setting scenarios: (1) Imbalanced Source Setting: The average size in source shared classes is set to five different cases, namely 10, 20, 50, 70 and 100 for each class,with five transfer tasks: $M300(\textbf{10})\to R50(60)$, $M300(\textbf{20})\to R50(60)$, $M300(\textbf{50})\to R50(60)$, $M300(\textbf{70})\to R50(60)$ and $M300(\textbf{100})\to R50(60)$. The values in parentheses indicate the average size in the shared class. (2) Imbalanced Target Setting: a) \textbf{Sparse scenario}, we adopt the \emph{Leave-Five-Out }cross-validation method to verify the sparsity of the target domain. For every shared class index $i=1,\ldots,n/5$ in target domain, set these 5 classes with index $i$ to sparse that contains only 10 samples, 60 samples in each of the other 45 non-sparse classes; then, calculate the accuracy and f1 value of 5 sparse class $i$ and 45 non-sparse classes respectively; at the end, average the test results:  $Acc_{(n)}=\frac{1}{n}\sum_{1}^{n}acc_{(i)}$, $F1_{(n)}=\frac{1}{n}\sum_{1}^{n}f1_{(i)}$. 	b) \textbf{Imbalanced scenario}, the first 25 classes are set to sparse, each with 30 images; the last 25 classes set to non-sparse, each with 60 images. Calculate and average the results of 25 sparse classes and 25 non-sparse classes respectively.

\textbf{Baselines.} We compare the proposed GADA with state-of-the-art PDA and adversarial UDA methods, including  BA$^3$US\cite{liang2020balanced}, ResNet\cite{he2016deep}, CDAN\cite{long2018conditional}, MDD\cite{Zhang2019}, GVB-GD\cite{cui2020gvb}, GCAN\cite{Ma2019}, TAN\cite{Xiao2020}, CDAN+TAN and SGR\cite{liang2018symbolic}.

\textbf{Implementation details.} We implement all deep methods in PyTorch, and the Imagenet-pre-trained ResNet-50 is used as backbone. Our HGR layer can be inserted between the last layer of CNN and the global pooling layer, which produces 2048 feature maps with $7 \times 7$ size. In the Office-Home dataset, the concept hierarchy graph with 87 symbolic nodes is generated by mapping 65 classes into WordTree, see in appendices. In the Meal-300 datasets, 300 classes are mapped to a conceptual hierarchy graph with 337 symbol nodes according to the provided hierarchical relationship\cite{Xiao2020}. Generally, $\gamma $ is set to 0.7. We adopt the SGD optimizer with learning rate 0.001, nesterov momentum 0.9, and weight decay 0.0005. 

\subsection{The Analysis of Experimental Results}

\begin{table*}[!t]
	\renewcommand{\arraystretch}{1.0}
	\small
	\caption{Classification accuracy and f1 value (\%) on Office-Home dataset, Imbalance setting (At least size).}
	\label{Tab-office-home}
	\centering
	\setlength{\tabcolsep}{0.7mm}{
		\begin{tabular}{c| c c c c c c c | c  c c  c c c c c c  }
			\hline
			 & \multirow{3}{*}{Method} & \multicolumn{6}{c}{Imbalanced Source Tasks} & \multicolumn{6}{c}{Imbalanced Target Tasks} & 
			 &&
			
			 \\
			  \cline{3-14}
			 
			 &            & \multicolumn{2}{c}{\textbf{Ar}$\rightarrow$\textbf{Cl}} & \multicolumn{2}{c}{\textbf{Ar}$\rightarrow$\textbf{Pr}} & \multicolumn{2}{c}{\textbf{Ar}$\rightarrow$\textbf{Rw}} & \multicolumn{2}{c}{\textbf{Cl}$\rightarrow$\textbf{Ar}} & \multicolumn{2}{c}{\textbf{Pr}$\rightarrow$\textbf{Ar}} & \multicolumn{2}{c}{\textbf{Rw}$\rightarrow$\textbf{Ar}} & \multirow{2}{*}{Avg-acc} & \multirow{2}{*}{Avg-f1}                                                                                                                   \\
			\cline{3-14}
			 &           & acc                                                     & f1                                                      & acc                                                     & f1                                                      & acc                                                     & f1                                                      & acc                      & f1                      & acc              & f1               & acc              & f1                                                     \\
			\hline
			\multirow{2}{*}{\shortstack{ PDA \\ $65\rightarrow$25}}
 			 & BA$^3$US & 37.39 & 17.36 &	67.43 &	30.01 &	66.20 & 28.23	& 53.62 &	29.15 &	59.00 &	27.17 &	63.14& 32.17 &	57.79 &	27.34         \\
			 \cline{2-16}
			 & BA$^{3}$US+HGR & 41.65 &	18.6 &	74.3 &	35.98 &	70.6 &	30.44 &	56.41 &	29.31 &	59.51 &	29.42 &	63.49 & 31.62 & 	60.99 & 	29.22 \\ 
			\hline
			\multirow{6}{*}{\shortstack{ UDA\\ $S25 \rightarrow$ \\ T25}}
			 & ResNet & 46.89 &	44.83 &	70.62 &	68.86&	74.80 &	73.18	&49.27 &	46.95 &	55.48 &	54.37 &	66.66 &	66.89&	60.62 &	59.18  \\
			 
			 & CDAN & 58.40 & 57.37 &	81.05 &	79.39&	82.4 &	81.19&	57.55&	57.37&	63.14&	62.96&	75.15&	74.73&		69.61&	68.83\\ 
			  & GVB-GD & 61.20 & 60.54 & 80.11 &	78.64 &	85.06 &	84.24 &	63.14 &	62.74	& 67.28	& 51.34	& 78.88	& 77.57 & 72.61 &	69.17  \\
			 & GCAN  & 62.42 &61.28 &	82.70 &	81.18 &	84.66 &	83.33 &	63.14 &	62.11 &	69.97 &	69.77 &	77.01 &	75.89	& 	73.31 &	72.26 \\

			 & MDD & 62.69	& 61.64 &83.37 &	81.52 &	85.06 &	83.73 &	62.28 &	61.95 &	67.26 &	66.32 &	77.13 &	76.95 &	72.96 &	72.01

\\
			
			 \cline{2-16}
			  &MDD+HGR & 63.88 & 63.02 &	85.11 &	83.64 &	85.53	& 84.34 &	63.56	 & 64.21 &	71.22 &	70.86 &77.65 &	77.32 &	74.49 &	73.89  \\

			\hline
			\multirow{6}{*}{\shortstack{ NI-UDA\\ $S65 \rightarrow$\\ $S25 \rightarrow$ \\ T25}}
			 & TAN  & 52.86 &	50.90 &	81.05 &	78.55 &	82.20 &	80.75 &	59.62 &	59.36 &	64.80 &	63.83 &	73.29 &	72.93 & 68.97 & 67.72                    \\
			 & CDAN+TAN & 59.25 &	58.63 &	83.64 &	81.61 &	83.46 &	82.27 &	59.83 &	60.04 &	65.04 &	64.85 &	74.12 &	73.33	&	70.89 &	70.12                    \\
			 \cline{2-16}
			& GADA(CDAN+SGR) & 59.92 &	59.71 &	83.47&	81.44&	83.8	&83.68&	59.00&	58.76&	65.42&	64.79&	76.31&	75.05&		71.32&	70.57			 \\ 
			& GADA(CDAN+HGR) & 61.57&	61.24&	84.45&	83.76&	84.76&	84.12&	60.9&	59.17	&65.76	&65.45&	77.01&	75.98&		72.40&	71.62 			\\
			& GADA(MDD+SGR) & 63.09	& 62.81	& 85.04 &	83.26 &	85.13	& 84.01 &	62.11 &	62.06 &	66.53 &	65.30 &	77.84&	77.46 &		73.29	&72.48 			 \\
			& GADA(MDD+HGR) & \bm{$66.87$} & \bm{$66.34$}	& \bm{$87.97$} &	\bm{$86.60$} &	\bm{$87.53$} &	\bm{$86.30$} &	\bm{$66.04$} &	\bm{$65.06$} &	\bm{$73.70$} &	\bm{$73.24$} &	78.46 &	78.11 &  \bm{$76.76$} & \bm{$75.94$}
	 \\ 
			& GADA(GVB+HGR)&  65.16 &	64.26 &	85.22 &	83.23 &	85.73	& 84.72 &	64.18 &	63.39 &	69.59 &	68.60 &	\bm{$79.08$} & \bm{$78.25$} &		74.82	& 73.74 \\ 
	
			\hline
		\end{tabular}}
\end{table*} 

\begin{table*}[!t]
	\renewcommand{\arraystretch}{1.0}
	\small
	\caption{Classification accuracy and f1 value (\%) on Office-Home datasets, Long-tail distribution( in  alphabetical order).}
	\label{Tab-office-home-alph}
	\centering
	\setlength{\tabcolsep}{1.2mm}{
		\begin{tabular}{ c c c c c c c | c  c c  c c c c c c  }
			\hline
			  \multirow{2}{*}{Method} & \multicolumn{2}{c}{\textbf{Ar}$\rightarrow$\textbf{Cl}} & \multicolumn{2}{c}{\textbf{Ar}$\rightarrow$\textbf{Pr}} & \multicolumn{2}{c}{\textbf{Ar}$\rightarrow$\textbf{Rw}} & \multicolumn{2}{c}{\textbf{Cl}$\rightarrow$\textbf{Ar}} & \multicolumn{2}{c}{\textbf{Pr}$\rightarrow$\textbf{Ar}} & \multicolumn{2}{c}{\textbf{Rw}$\rightarrow$\textbf{Ar}} & \multirow{2}{*}{Avg-acc} & \multirow{2}{*}{Avg-f1}                                                                                                                   \\
		\cline{2-13}
        & acc & f1 & acc & f1   & acc & f1   & acc & f1   & acc & f1   & acc & f1  \\
			\hline
		
			  BA3$^3$US\cite{liang2020balanced}  & 60.62 & 26.50 & 83.16 & 45.73& 88.39 &	54.71 & 71.75& 34.84&75.45&35.21&79.25	&39.21 & 	76.43 & 39.36          \\
		
			\hline
			
			MDD & 71.16	& 70.18	& 90.51 & 90.12 &	91.32 &	90.37&	79.4	&74.22&	79.49&	75.74&	85.19&	82.52	&	82.84 &	80.52 \\

			\hline

			 GADA(MDD+HGR)& 73.11 &	72.48 & 92.13 & 91.46 &	93.1 &	91.79	& 81.81 &	76.28 &	82.78 &	78.23 &	86.42 &	83.67	&84.89	& 82.31 \\

			\hline
		\end{tabular}}
\end{table*}

\begin{table*}[!t]
	\renewcommand{\arraystretch}{1.0}
	\small
	\caption{Accuracy (\%) on Office-Home for unsupervised domain adaptation(S65 $\to$ T65).}
	\label{Tab-office-home-uda65}
	\centering
	\setlength{\tabcolsep}{0.6mm}{
		\begin{tabular}{ c c c c c c c | c  c c  c c c c   }
			\hline
			  \multirow{1}{*}{Method} & \multicolumn{1}{c}{\textbf{Ar}$\rightarrow$\textbf{Cl}} & \multicolumn{1}{c}{\textbf{Ar}$\rightarrow$\textbf{Pr}} & \multicolumn{1}{c}{\textbf{Ar}$\rightarrow$\textbf{Rw}} & \multicolumn{1}{c}{\textbf{Cl}$\rightarrow$\textbf{Ar}} & \multicolumn{1}{c}{\textbf{Cl}$\rightarrow$\textbf{Pr}} & \multicolumn{1}{c}{\textbf{Cl}$\rightarrow$\textbf{Rw}} &
			  \multicolumn{1}{c}{\textbf{Pr}$\rightarrow$\textbf{Ar}} &
			  \multicolumn{1}{c}{\textbf{Pr}$\rightarrow$\textbf{Cl}} &
			  \multicolumn{1}{c}{\textbf{Pr}$\rightarrow$\textbf{Rw}} &
			  \multicolumn{1}{c}{\textbf{Rw}$\rightarrow$\textbf{Ar}} &
			  \multicolumn{1}{c}{\textbf{Rw}$\rightarrow$\textbf{Cl}} &
			  \multicolumn{1}{c}{\textbf{Rw}$\rightarrow$\textbf{Pr}} &
			  \multirow{1}{*}{Avg-acc} \\                                                                                                              
			\hline
		
			MDD\cite{Zhang2019} & 54.9 & 73.7 & 77.8 & 60.0 & 71.4 & 71.8 & 61.2 & 53.6 & 78.1 & 72.5 & 60.2 & 82.3 & 68.1 \\

			\hline
			 GADA(MDD+HGR)&   56.4 & 74.8 & 79.2 & 61.5 & 72.8 & 73.1 & 62.5 & 54.8 & 79.2 & 73.7 & 61.4 & 83.6 & 69.4  \\

			\hline
		\end{tabular}}
\end{table*}

\begin{table}[!t]
	\renewcommand{\arraystretch}{1.0}
	\small
	\caption{Classification f1 value (\%) on Meal-300 dataset in Imbalanced Target Setting. (A(60): 45 non-sparse classes, A(10): 5 sparse classes; B(60): 25 non-sparse classes, B(30): 25 sparse classes; C(10): all sparse)}
	\label{Tab-meal300-target}
	\centering
	\setlength{\tabcolsep}{0.7mm}{
		\begin{tabular}{c|ccc|cc|c}
				\hline
				& \multirow{2}{*}{Method} & \multicolumn{1}{c}{A(60) } & \multicolumn{1}{c}{A(10)} & \multicolumn{1}{c}{B(60)} & \multicolumn{1}{c}{B(30)}   & \multicolumn{1}{c}{C(10)}                                                                         \\
				\cline{3-7}
				&       & f1      & f1 & f1 & f1    & f1    \\
				\hline
				\multirow{2}{*}{\shortstack{ PDA\\ 300$\rightarrow$50}}
			
				& BA3US   &   40.91& 15.09 & 25.16    & 25.92    &  21.23                                                           \\ 
				\cline{2-7}
				& BA$^{3}$US+HGR & 41.22 & 19.48 & 25.50 & 26.02 & 23.17 \\ 
				\hline
				
				\multirow{6}{*}{\shortstack{ UDA\\ $M50 \rightarrow$ \\ R50 }}
				& ResNet                                           & 52.73                                                 & 12.61                                      & 30.20                     & 37.14      & 59.07    \\
				
				& CDAN                                       & 55.52                                              & 12.19                                    & 31.57                        & 37.13       & 71.28   \\
				& GVB-GD & 	69.13 &	30.04 &	43.98 & 47.64  & 75.18 \\
				& MDD   &	65.42 & 21.96 & 40.13 &	44.93  & 72.16 		    \\ 
			
				\cline{2-7}
				& MDD+HGR &	70.32 & 28.72 & 41.65 &	45.36 & 73.11  \\
				& GVB+HGR &	69.82 & 30.74 & 44.25 &	48.36 & 75.89  \\
				\hline
				
				\multirow{4}{*}{\shortstack{ NI-UDA\\ M300 \\ $\rightarrow$   R50 }}
				& TAN                  & 53.41                      & 13.60                   & 30.73                     & 38.22   		& 	60.77 	\\
				\cline{2-7}
				& GADA(MDD+SGR)   &	67.35 	&22.35&	40.29 &	46.01 & 73.80 \\ 
				& GADA(MDD+HGR)  & \bm{$72.10$}  &	30.84 &	43.31  &	47.23 & 76.08 \\
				& GADA(GVB+HGR)	 & 70.67  &	\bm{$32.89$} &	\bm{$44.75$}  &	\bm{$50.90$} & \bm{$77.21$} \\
				\hline
		\end{tabular}}
\end{table}

\textbf{Meal-300.} Table \ref{Tab-meal300-source}  and  \ref{Tab-meal300-target} report the detailed comparison of two imbalance settings in NI-UDA on Meal-300 datasets. 

(1) \textbf{Imbalanced source scenario}, Our GADA(HGR) improves f1 of the most adversarial UDA methods on average by  \textbf{4\%}(MDD) and \textbf{3.79\%}(GVB-GD). Specifically, in the case of high sparsity factor, e.g. $Meal300(10) \rightarrow Real50(60)$, GADA(HGR) can greatly improve transfer performance, \textbf{7.19\% }is improved on MDD, and \textbf{7.89\%} is improved  on GVB-GD. In near-balance case, e.g. $Meal300(100) \rightarrow Real50(60)$, GADA(HGR) can still get significantly improvements such as \textbf{3.31\%} and \textbf{2.95\%} points improvement respectively on MDD and GVB-GD. It shows our GADA framework can make full use of non-shared 
classes knowledge through hierarchical graph reasoning to solve two challenges of NI-UDA in imbalanced source scenario.

(2) \textbf{Imbalanced target scenario}, GADA(MDD+HGR) improves 1) the f1 value of \textbf{6.68\%} and \textbf{8.88\%} at 45 non-sparse classes and 5 sparse classes, respectively; 2) the f1 value of \textbf{3.18\%} and \textbf{2.3\%} at 25 non-sparse and 25-sparse classes, respectively; 3) \textbf{3.92\% }f1 in the fully sparse target domain, where all the target classes are sparse with 10 samples per class. GADA(GVB-GD+HGR) also has consistency improvement on these three cases, see Table \ref{Tab-meal300-target}. The results demonstrate that GADA can also solve two challenges of NI-UDA in imbalanced target scenario well.

\textbf{Office-Home.} Table  \ref{Tab-office-home} shows the comparison results for 3 imbalanced source tasks and 3 imbalanced target tasks under the imbalance setting of Office-home datasets. On average, our GADA(HGR) improves f1 of the most adversarial UDA methods by \textbf{2.79}\%(CDAN) and \textbf{3.93\%}(MDD) and \textbf{4.57\%}(GVB-GD). The experimental results show that our GADA framework can well solve two challenges of NI-UDA in both imbalanced source and target scenarios.


To compare PDA methods, we conduct the experiments on origin PDA setting ($S65\to T25$)\cite{liang2020balanced}, where first 25 classes in alphabetical order are included as shared classes, which is a long-tailed distribution. The comparison results are shown in Table \ref{Tab-office-home-alph}. The BA$^3$US of Imbalanced setting is degraded in compare with PDA setting, it shows Office-Home under imbalanced setting is more challenging and consistent with the NI-UDA problem.  GADA still has consistent improvement. 

To compare UDA methods, we also conduct the experiments on origin UDA  ($S65\to T65$)\cite{Zhang2019}. Table \ref{Tab-office-home-uda65} shows the results of 65 classes in near-balanced UDA settings. GADA still achieves a competitive improvement, which proves the graph reasoning ability of HGR.

In summary, GADA in NI-UDA has achieved a significantly improvement on the Office-Home dataset with a little semantic relatedness. However, GADA's improvement will become large margin on the Meal-300 datasets which has high semantic relatedness. GADA(HGR) shows its superiority power in the aggregation of non-shared knowledge from big data.


\subsection{Ablation Study}  
 
\begin{table}[!t]
\renewcommand{\arraystretch}{1.1}
\small
\caption{GADA Ablation Study}
\label{Tab-gada-ablation}
\centering
\begin{tabular}{c| c  c c c c  c  }
\hline
& \multirow{2}{*}{Methods}  & \multicolumn{2}{c}{M300(10) $\rightarrow$ R50}   & \multicolumn{2}{c}{Ar $\rightarrow$ Cl}      \\
\cline{3-6}        
& &  acc    & f1     &  acc              & f1          \\
\hline 
UDA  & MDD & 64.13  & 64.42 & 62.69	& 61.64\\ 
\hline
\multirow{8}{*}{\shortstack{ NI-UDA \\ GADA}}
& Ours (Baseline) & 68.86 & 68.54 & 63.09 & 62.81 \\  
& Ours (w/ BN)   & 69.79 & 69.46 & 64.27 & 63.83\\
& Ours (w/ MLP)  & 70.43 & 70.17 & 64.80 & 64.54 \\ 
& Ours (w/ HAP)  & 70.94 & 70.50 & 65.45 & 65.16\\ 
\cline{2-6}
& Ours (w/ HGR)& 71.92 & 71.62 & 66.87 & 66.34 \\
\cline{2-6}
&$L_K^1 \qquad L_K^2$ & \\
\cline{2-6}
 & $\checkmark \qquad \quad $ & 71.23 & 71.05 & 66.13 & 65.66 \\
& $\checkmark \qquad \checkmark $ & 71.92 & 71.62 & 66.87 & 66.34 \\
\hline
\end{tabular}

\end{table}

In order to exploit the components of HGR and GADA contribute to the performance, we perform  ablation studies on imbalanced  task of  $Meal300(10) \rightarrow Real50(60), Art  \to  Clipart $. SGR\cite{liang2018symbolic} is symbolic graph reasoning with feature enhance for classification task and semantic segmentation task. We also propose this general graph reasoning SGR for classification task as a baseline hierarchy graph reasoning component of GADA in NI-UDA. So, we regard the "GADA(MDD+SGR)" model as the baseline. We study the influence of HAP(Hierarchy Attention with Prediction), MLP(Multilayer Perception Mapping), and BN(Batch Normalization) of HGR as show in Table \ref{Tab-gada-ablation}.  
 
In Meal-300 datasets, HAP can promote the baseline by around 1.96\% f1. MLP achieves around 1.63\% f1 improvements. BN apparently boosts the f1 by around 0.92\% from baseline. GADA(MDD+HGR) improves baseline by around 3.08\%. In Office-Home dataset, there is still consistency improvement of imbalanced source task. In compare with baseline GADA(MDD+SGR) in NI-UDA, our method GADA(MDD+HGR) has significantly improvement which shows HGR can borrow the knowledge from non-shared big data sufficiently.

To study the influence of the source domain confidence filtering mechanism of GADA, $L_K^1$ and $L_K^2$ of GADA ablation analysis are demonstrated in last three lines of Table \ref{Tab-gada-ablation}. It shows that the Source Classifier Filter (SCF) mechanism can alleviate the negative transfer effect caused by non-shared classes. 

In compare with MDD in UDA, our method GADA(MDD+HGR) has superiority performance in NI-UDA which show non-shared data is useful for domain adaptation and our GADA is sufficient to use these non-shared data while solved the negative transfer challenge of NI-UDA.



\subsection{Parameter sensitivity analysis}

	

\textbf{Hyper parameters of $\lambda_1,\lambda_2$.}
We investigate a broader scope of our algorithm GADA for source classier and domain discriminator by varying parameters $\lambda_1,\lambda_2$. When $\lambda_1$ is 0, NI-UDA settings will be converted to UDA settings. Generally, the $ (\lambda_1,\lambda_2)$ of GADA(MDD+HGR) on the Office-home and Meal300 datasets are (2, 4) and (2, 3.2) respectively.



\begin{table}[h] 
	\renewcommand{\arraystretch}{1.2}
	\small
	\caption{The hyper-parameters of ${\lambda}_1$ ,$ {\lambda}_2$   in the overall loss of GADA} 
	\label{Tab-hyper-lambda}
	\centering  
	\setlength{\tabcolsep}{2mm}{
		\begin{tabular}{ c c c c c c  }
			\hline
			\multirow{2}{*}{$ {\lambda}_1$ } &  \multirow{2}{*}{$ {\lambda}_2$ }  & \multicolumn{2}{c}{M300(10) $\to$ R50} & \multicolumn{2}{c}{Ar $\to$ Cl} \\
			\cline{3-6}
			 &  & acc & f1  & acc & f1 \\
			\hline
	         0 & 1 & 65.87 &	66.09 &63.88 &	63.02 \\
	         2 & 4 & 71.92 & 71.62 & 66.54 & 65.89\\ 
	         2 & 3.2 & 71.34 & 70.92 & 66.87 & 66.34 \\
			\hline
	\end{tabular}}

\end{table}

\textbf{The dimension of  $D^S$.} 
we transform the local feature space to the semantic space with a two layer MLP as :\begin{small}
	\begin{equation}
	X^S = MLP(X^l)  
	\end{equation}
\end{small}where $X^S\in \mathbb{R}^{H^l\times W^l\times D^S}$ and $D^S$ is the designed size of semantic nodes, i.e., 2048 $\underrightarrow{1 \times1 \, conv}$ 512 $\underrightarrow{1 \times1 \, conv}$ 128. We also have semantic nodes $S^a$ and visual representations $H^{s}$ belongs to $\mathbb{R}^{|\mathcal{N}| \times D^s}$. We study the influence of different sizes of $D^S$, like 256-d and 128-d, in transfer task of $Meal300(10) \to Real(60)$ on GADA(MDD+HGR). \par 
As shown in Table \ref{Tab-hyper}, compared with 128-d, 256-d get a slight improvement and more parameters. In fact, we set the dimension of $D^S$ to 128, in order to reduce the training parameters.

\begin{table}[h] 
	\renewcommand{\arraystretch}{1.2}
	\small
	\caption{Parameter sensitivity analysis of $D^S$ and num of HGR layer} 
	\label{Tab-hyper}
	\centering  
	\setlength{\tabcolsep}{1mm}{
		\begin{tabular}{c | c c c c }
			\hline
			\multirow{2}{*}{\shortstack{num of \\ HGR layer}} &	\multirow{2}{*}{\shortstack{$X^S\in$ \\ $ \mathbb{R}^{H^l\times W^l\times D^S}$}} &  \multirow{2}{*}{\shortstack{ $H^S\in $ \\ $\mathbb{R}^{\mathcal{N} \times D^S}$}}  & \multicolumn{2}{c}{M300(10) $\to$ R50}   \\
			\cline{4-5}
			
			& &   & acc & f1  \\
			\hline
			1 & 128 & 128 &  71.92 & 71.62 \\ 
			1 & 256 & 256 &  72.19 & 71.80  \\ 
			2 & 128 & 128 &  70.73 & 70.59 \\
			\hline
	\end{tabular}}

\end{table}

\textbf{Nums of HGR layer.}
 We also studied the influence of different layers of HGR layer. Table \ref{Tab-hyper} show the result: one layer of HGR has better effect than two layers. \par 
 
 \subsection{Visualization}   
\begin{figure}[t]  
	\subfigure[ MDD] { 
		\begin{minipage}[t]{0.48\linewidth} 
			\centering   
			\includegraphics[width=1.72in]{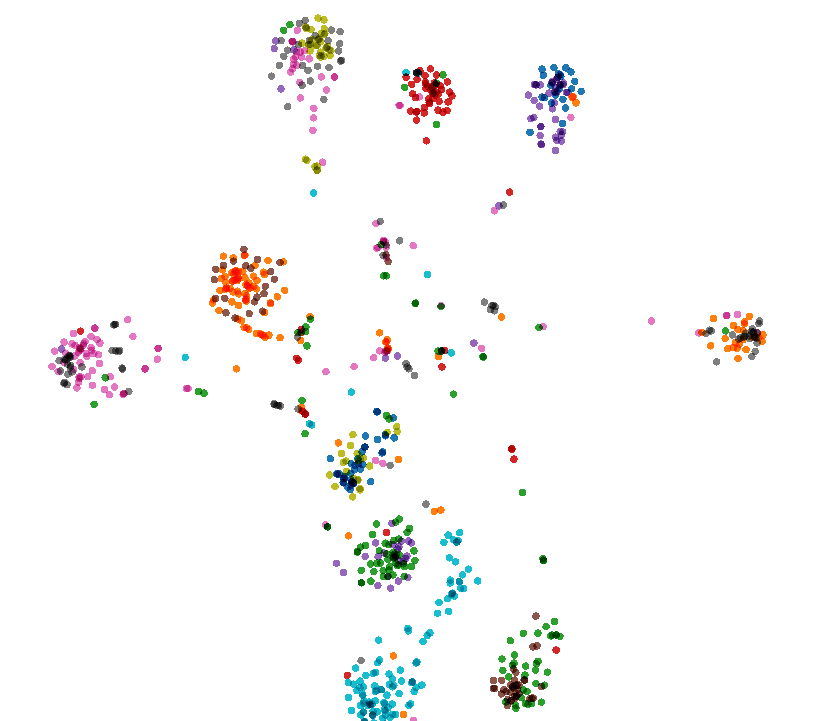}    
	\end{minipage}}
	\subfigure[ GADA(MDD+HGR)]{ 
		\begin{minipage}[t]{0.48\linewidth} 
			\centering   
			\includegraphics[width=2.22in]{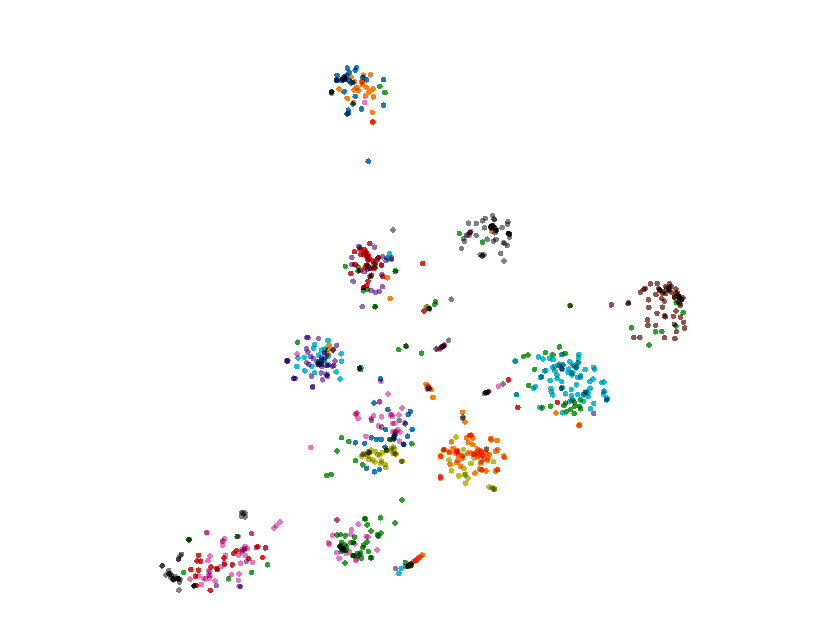}    
	\end{minipage}}

	\caption{t-SNE visualization of imbalanced tasks Ar $\to$ Cl. } 
	\label{fig:tsne} 
\end{figure}

 We utilize t-SNE to visualize  the feature distributions of Imbalanced tasks Ar $\to$ Cl  on Office-Home datasets(10 classes). As shown in Fig. \ref{fig:tsne}, the imbalanced scenarios of either domain will cause NP, NF problems, which manifests as some samples drifting near the center point of domain invariant clustering. Obviously, GADA model has a better clustering effect. More visualization can be seen in the appendix.
 
 \subsection{Discussion of GADA}  
For PDA methods, including SOTAs of ETN\cite{cao2019learning} and BA$^3$US\cite{liang2020balanced} do not analyze the f1 value which is important for our imbalanced setting. From the experimental results, we can draw conclusions: Under the imbalanced setting of PDA, the f1 value will collapse and decrease,and cause a lot of NF, NP problems. 

In general, Our GADA has achieved a huge improvement in NI-UDA. Surprisingly, HGR can consistently improve the SOTAs of PDA and UDA. Because GADA can leverage priori hierarchy knowledge to enhance domain adversarial aligned and class structure aligned feature representation. In addition, most UDA methods can be easily extended to NI-UDA.  GADA with HGR can consistently improve the SOTAs models of PDA, UDA and NI-UDA, which can be regarded as a general adversarial alignment paradigm. In the future work, we will adopt data augmentation, consistency training, self-training, etc. to further alleviate the challenges of NI-UDA.

\section{Conclusion}
In this paper, we present a novel approach (GADA model) for the  non-shared-and-imbalanced domain adaptation. The scope of this approach is the assumption of existing sibling relationship between shared classes and non-shared classes. For knowledge sharing from non-shared data without negative transfer effect challenge, our SCF could filter non-shared data with high-confidence non-shared data in feature enhance layer. For sparse classes transfer challenge, HGR could enhance the sibling aligned feature representation for adversarial domain adaptation. We also apply GADA to existing methods and achieve remarkable improvement over original counterparts.

\appendices
\section{Hierarchical graph construction }
\begin{figure}[!h]  
	\begin{minipage}[b]{1\linewidth} 
		\centering   
		\includegraphics[width=3.4in]{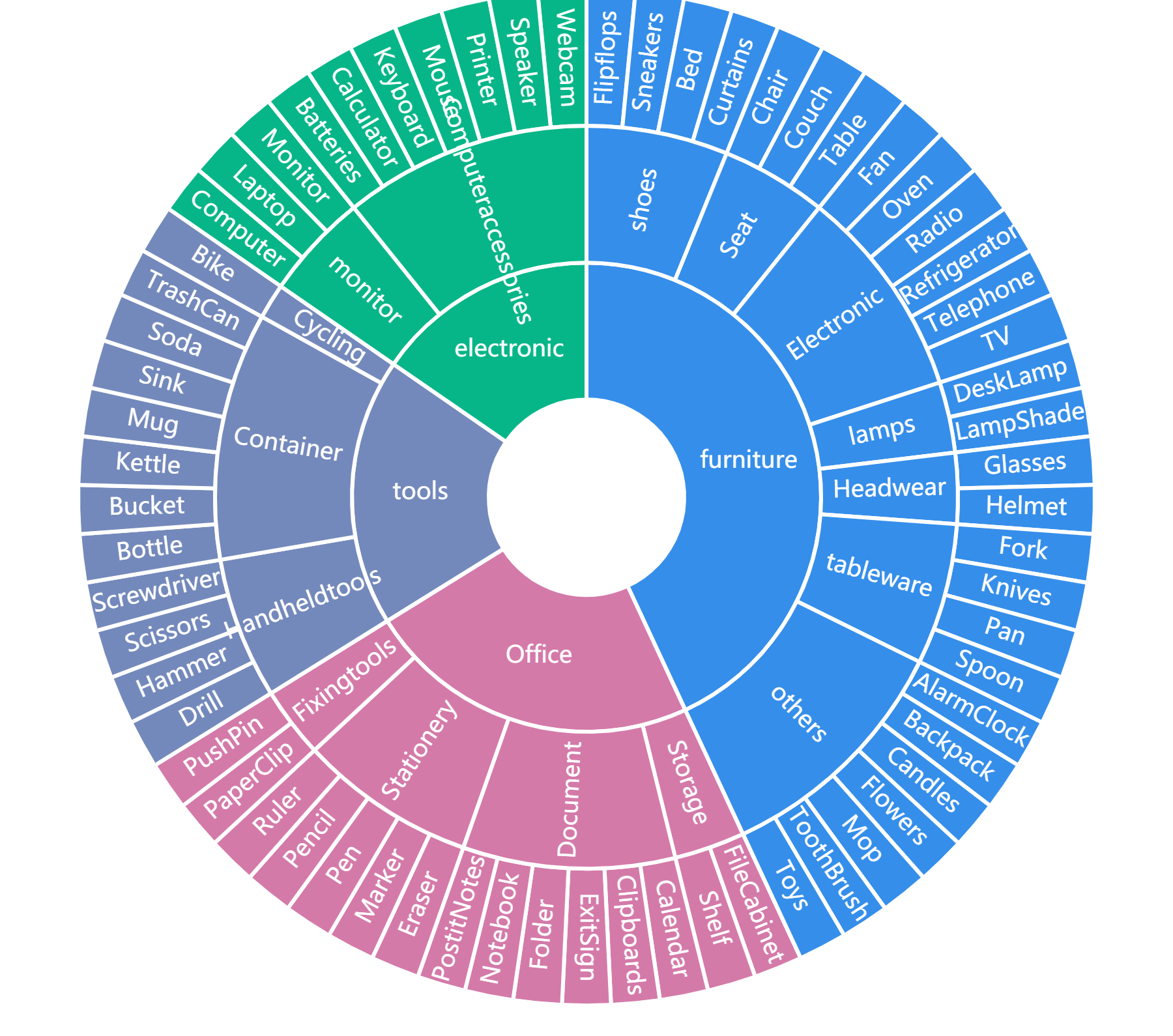}    
		\caption{A hierarchical graph of 87 nodes constructed by the office-home datasets. 	}
		\label{office_graph_65}
	\end{minipage}%
	\hfill  
\end{figure}

The hierarchical graph constructed according to WordTree is shown in Figure \ref{office_graph_65}.




\ifCLASSOPTIONcaptionsoff
  \newpage
\fi

\bibliographystyle{IEEEtran}
\bibliography{tnnls21}

\begin{thebibliography}{10}
\providecommand{\url}[1]{#1}
\csname url@samestyle\endcsname
\providecommand{\newblock}{\relax}
\providecommand{\bibinfo}[2]{#2}
\providecommand{\BIBentrySTDinterwordspacing}{\spaceskip=0pt\relax}
\providecommand{\BIBentryALTinterwordstretchfactor}{4}
\providecommand{\BIBentryALTinterwordspacing}{\spaceskip=\fontdimen2\font plus
\BIBentryALTinterwordstretchfactor\fontdimen3\font minus
  \fontdimen4\font\relax}
\providecommand{\BIBforeignlanguage}[2]{{%
\expandafter\ifx\csname l@#1\endcsname\relax
\typeout{** WARNING: IEEEtran.bst: No hyphenation pattern has been}%
\typeout{** loaded for the language `#1'. Using the pattern for}%
\typeout{** the default language instead.}%
\else
\language=\csname l@#1\endcsname
\fi
#2}}
\providecommand{\BIBdecl}{\relax}
\BIBdecl

\bibitem{zhuang2019comprehensive}
F.~Zhuang, Z.~Qi, K.~Duan, D.~Xi, Y.~Zhu, H.~Zhu, H.~Xiong, and Q.~He, ``A
  comprehensive survey on transfer learning,'' \emph{arXiv preprint
  arXiv:1911.02685}, 2019.

\bibitem{yosinski2014transferable}
J.~Yosinski, J.~Clune, Y.~Bengio, and H.~Lipson, ``How transferable are
  features in deep neural networks?'' in \emph{Advances in neural information
  processing systems}, 2014, pp. 3320--3328.

\bibitem{he2016deep}
K.~He, X.~Zhang, S.~Ren, and J.~Sun, ``Deep residual learning for image
  recognition,'' in \emph{Proceedings of the IEEE conference on computer vision
  and pattern recognition}, 2016, pp. 770--778.

\bibitem{ding2018graph}
Z.~Ding, S.~Li, M.~Shao, and Y.~Fu, ``Graph adaptive knowledge transfer for
  unsupervised domain adaptation,'' in \emph{Proceedings of the European
  Conference on Computer Vision (ECCV)}, 2018, pp. 37--52.

\bibitem{cui2020gradually}
S.~Cui, S.~Wang, J.~Zhuo, C.~Su, Q.~Huang, and Q.~Tian, ``Gradually vanishing
  bridge for adversarial domain adaptation,'' in \emph{Proceedings of the
  IEEE/CVF Conference on Computer Vision and Pattern Recognition}, 2020, pp.
  12\,455--12\,464.

\bibitem{chang2019domain}
W.-G. Chang, T.~You, S.~Seo, S.~Kwak, and B.~Han, ``Domain-specific batch
  normalization for unsupervised domain adaptation,'' in \emph{Proceedings of
  the IEEE Conference on Computer Vision and Pattern Recognition}, 2019, pp.
  7354--7362.

\bibitem{cao2019learning}
Z.~Cao, K.~You, M.~Long, J.~Wang, and Q.~Yang, ``Learning to transfer examples
  for partial domain adaptation,'' in \emph{Proceedings of the IEEE Conference
  on Computer Vision and Pattern Recognition}, 2019, pp. 2985--2994.

\bibitem{long2018conditional}
M.~Long, Z.~Cao, J.~Wang, and M.~I. Jordan, ``Conditional adversarial domain
  adaptation,'' in \emph{Advances in Neural Information Processing Systems},
  2018, pp. 1640--1650.

\bibitem{Zhang2019}
Y.~Zhang, T.~Liu, M.~Long, and M.~Jordan, ``Bridging theory and algorithm for
  domain adaptation,'' in \emph{International Conference on Machine Learning},
  2019, pp. 7404--7413.

\bibitem{Xiao2020}
G.~{Xiao}, Q.~{Wu}, H.~{Chen}, D.~{Cao}, J.~{Guo}, and Z.~{Gong}, ``A deep
  transfer learning solution for food material recognition using electronic
  scales,'' \emph{IEEE Transactions on Industrial Informatics}, vol.~16, no.~4,
  pp. 2290--2300, 2020.

\bibitem{Chen2019}
C.~Chen, W.~Xie, W.~Huang, Y.~Rong, X.~Ding, Y.~Huang, T.~Xu, and J.~Huang,
  ``Progressive feature alignment for unsupervised domain adaptation,'' in
  \emph{Proceedings of the IEEE Conference on Computer Vision and Pattern
  Recognition}, 2019, pp. 627--636.

\bibitem{liang2020balanced}
J.~Liang, Y.~Wang, D.~Hu, R.~He, and J.~Feng, ``A balanced and
  uncertainty-aware approach for partial domain adaptation,'' \emph{arXiv
  preprint arXiv:2003.02541}, 2020.

\bibitem{miller1998wordnet}
G.~A. Miller, \emph{WordNet: An electronic lexical database}.\hskip 1em plus
  0.5em minus 0.4em\relax MIT press, 1998.

\bibitem{yan2017mind}
H.~Yan, Y.~Ding, P.~Li, Q.~Wang, Y.~Xu, and W.~Zuo, ``Mind the class weight
  bias: Weighted maximum mean discrepancy for unsupervised domain adaptation,''
  in \emph{Proceedings of the IEEE Conference on Computer Vision and Pattern
  Recognition}, 2017, pp. 2272--2281.

\bibitem{li2018adaptive}
Y.~Li, N.~Wang, J.~Shi, X.~Hou, and J.~Liu, ``Adaptive batch normalization for
  practical domain adaptation,'' \emph{Pattern Recognition}, vol.~80, pp.
  109--117, 2018.

\bibitem{jiang2020implicit}
X.~Jiang, Q.~Lao, S.~Matwin, and M.~Havaei, ``Implicit class-conditioned domain
  alignment for unsupervised domain adaptation,'' \emph{arXiv preprint
  arXiv:2006.04996}, 2020.

\bibitem{liang2020we}
J.~Liang, D.~Hu, and J.~Feng, ``Do we really need to access the source data?
  source hypothesis transfer for unsupervised domain adaptation,'' \emph{arXiv
  preprint arXiv:2002.08546}, 2020.

\bibitem{cao2018pada}
Z.~Cao, L.~Ma, M.~Long, and J.~Wang, ``Partial adversarial domain adaptation,''
  in \emph{Proceedings of the European Conference on Computer Vision (ECCV)},
  2018, pp. 135--150.

\bibitem{panareda2017open}
P.~Panareda~Busto and J.~Gall, ``Open set domain adaptation,'' in
  \emph{Proceedings of the IEEE International Conference on Computer Vision},
  2017, pp. 754--763.

\bibitem{zhao2019learning}
H.~Zhao, R.~T.~d. Combes, K.~Zhang, and G.~J. Gordon, ``On learning invariant
  representation for domain adaptation,'' \emph{arXiv preprint
  arXiv:1901.09453}, 2019.

\bibitem{tan2019generalized}
S.~Tan, X.~Peng, and K.~Saenko, ``Generalized domain adaptation with covariate
  and label shift co-alignment,'' \emph{arXiv preprint arXiv:1910.10320}, 2019.

\bibitem{zhou2020graph}
J.~Zhou, G.~Cui, S.~Hu, Z.~Zhang, C.~Yang, Z.~Liu, L.~Wang, C.~Li, and M.~Sun,
  ``Graph neural networks: A review of methods and applications,'' \emph{AI
  Open}, vol.~1, pp. 57--81, 2020.

\bibitem{kipf2016semi}
T.~N. Kipf and M.~Welling, ``Semi-supervised classification with graph
  convolutional networks,'' \emph{arXiv preprint arXiv:1609.02907}, 2016.

\bibitem{chien2021adaptive}
E.~Chien, J.~Peng, P.~Li, and O.~Milenkovic, ``Adaptive universal generalized
  pagerank graph neural network,'' in \emph{International Conference on
  Learning Representations. https://openreview. net/forum}, 2021.

\bibitem{liang2018symbolic}
X.~Liang, Z.~Hu, H.~Zhang, L.~Lin, and E.~P. Xing, ``Symbolic graph reasoning
  meets convolutions,'' in \emph{Advances in Neural Information Processing
  Systems}, 2018, pp. 1853--1863.

\bibitem{Ma2019}
X.~Ma, T.~Zhang, and C.~Xu, ``Gcan: Graph convolutional adversarial network for
  unsupervised domain adaptation,'' in \emph{The IEEE Conference on Computer
  Vision and Pattern Recognition (CVPR)}, June 2019.

\bibitem{wang2020learning}
H.~Wang, M.~Xu, B.~Ni, and W.~Zhang, ``Learning to combine: Knowledge
  aggregation for multi-source domain adaptation,'' in \emph{European
  Conference on Computer Vision}.\hskip 1em plus 0.5em minus 0.4em\relax
  Springer, 2020, pp. 727--744.

\bibitem{ben2010theory}
S.~Ben-David, J.~Blitzer, K.~Crammer, A.~Kulesza, F.~Pereira, and J.~W.
  Vaughan, ``A theory of learning from different domains,'' \emph{Machine
  learning}, vol.~79, no. 1-2, pp. 151--175, 2010.

\bibitem{cui2020gvb}
S.~Cui, S.~Wang, J.~Zhuo, C.~Su, Q.~Huang, and T.~Qi, ``Gradually vanishing
  bridge for adversarial domain adaptation,'' in \emph{Proceedings of the IEEE
  Conference on Computer Vision and Pattern Recognition}, 2020.

\bibitem{lin2013network}
M.~Lin, Q.~Chen, and S.~Yan, ``Network in network,'' \emph{arXiv preprint
  arXiv:1312.4400}, 2013.

\bibitem{venkateswara2017deep}
H.~Venkateswara, J.~Eusebio, S.~Chakraborty, and S.~Panchanathan, ``Deep
  hashing network for unsupervised domain adaptation,'' in \emph{Proceedings of
  the IEEE Conference on Computer Vision and Pattern Recognition}, 2017, pp.
  5018--5027.

\end{thebibliography}

%
\begin{IEEEbiography}[{\includegraphics[width=1in,height=1.25in,clip,keepaspectratio]{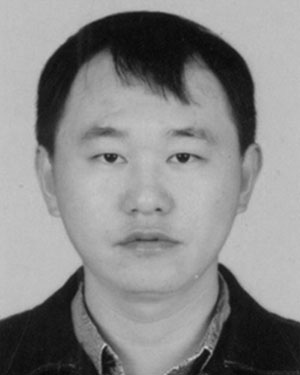}}]{Guangyi Xiao} (M’10) received the B.Econ. degree from Hunan University, Changsha, China, in 2006, the M.Sc. and Ph.D. degrees from the University of Macau, Macau, China, in 2009 and 2015, respectively, all in software engineering.

He is currently an Associate Professor in Colleague of Computer Science and Electronic Engineering with Hunan University. His current research interests include partial-and-imbalanced transfer learning, computer vision, semantic representation, semantic integration, semantic interoperation, and collaboration systems, mainly applied to the fields of food-computing, law-computing, e-commerce, and e-marketplace.

\end{IEEEbiography}

\begin{IEEEbiography}[{\includegraphics[width=1in,height=1.25in,clip,keepaspectratio]{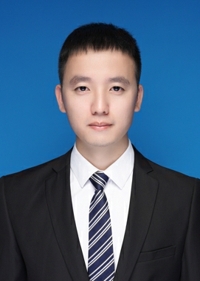}}]{Weiwei Xiang} received the B.Econ. degree in Electronic Information Engineering at Communication University of China, Beijing, China, in 2017, and
 M.S. degree in software engineering with the Hunan University, Changsha,China. His principal research is in the field of deep learning, computer vision and domain adaptation.
\end{IEEEbiography}

\begin{IEEEbiography}[{\includegraphics[width=1in,height=1.25in,clip,keepaspectratio]{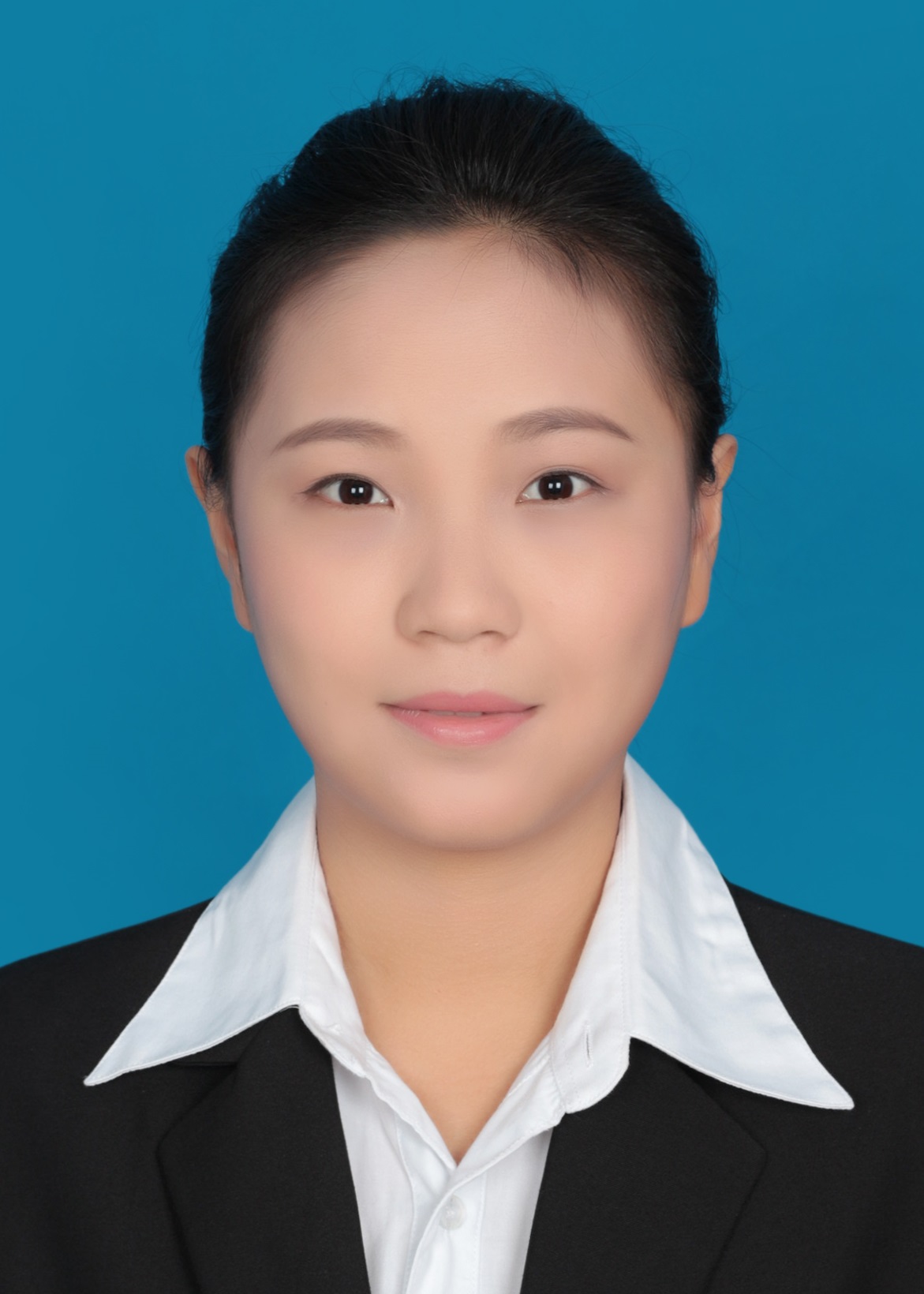}}]{Huan Liu} received the B.Econ. degree Computer Science and Technology in Xiangtan University, China, in 2017.
and M.S. degree  in computer science technology with the Hunan University, Changsha,China. His principal research is in the field of transfer learning.

\end{IEEEbiography}

\begin{IEEEbiography}[{\includegraphics[width=1in,height=1.25in,clip,keepaspectratio]{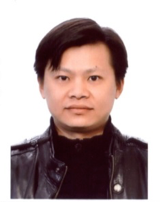}}]{Hao Chen} received the M.S. degree in software engineering from Hunan University, Changsha, China, in 2004, and the Ph.D. degree from the School of Information Science and Engineering, Changsha, China, in 2012. He is currently an Associate Professor and Ph.D. Supervisor at the College of Computer Science and Electronic Engineering, Hunan University, Changsha,China. His research interests are Web mining,personalized recommendation and big data technology.
\end{IEEEbiography}

\begin{IEEEbiography}[{\includegraphics[width=1in,height=1.25in,clip,keepaspectratio]{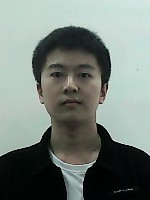}}]{Shun Peng} received the B.Econ. degree in Communication Engineering at Hunan Institute of Engineering, XiangTan, China, in 2020. He is currently a graduate student in Electronic information with the Hunan Univerisity, Changsha, China. His research focused on domain adaption, computer vision and deep learning.
 
\end{IEEEbiography}

\begin{IEEEbiography}[{\includegraphics[width=1in,height=1.25in,clip,keepaspectratio]{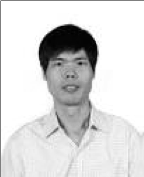}}]{Jingzhi Guo} received his PhD degree in Internet computing and e-commerce from Griffith University, Australia in 2005, the MSc degree in computation from the University of Manchester, UK, and the BEcon degree in international business management
from the University of International Business and Economics, China in 1988.
He is currently an Associate Professor in ecommerce technology with University of Macau, Macao. His principal researches are in the fields of semantic integration, virtual world and e-commerce technology.
\end{IEEEbiography}

\begin{IEEEbiography}[{\includegraphics[width=1in,height=1.25in,clip,keepaspectratio]{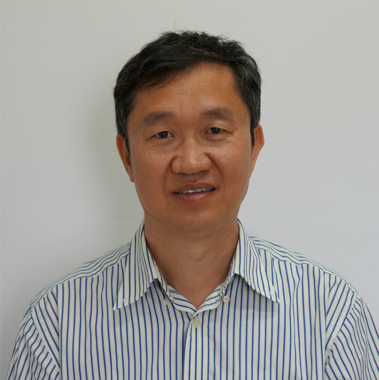}}]{Zhiguo Gong} received the PhD degree in the Department of Computer Science, Institute of Mathematics,Chinese Academy of Science, and the MSc degree from Peking University, Beijing, China, in 1988.
He is currently an Professor and the Head in the Department of Computer and Information Science,University of Macau, Macau, China. His research interests include Machine  Learning, Data Mining,Database, and Information Retrieval. He is a senior member of the IEEE.
\end{IEEEbiography}
\end{document}